\documentclass[a4paper,fleqn]{cas-dc}
    
    \usepackage[numbers]{natbib}
    \usepackage{amsmath}
    \usepackage{amssymb}
    \usepackage{booktabs}
    \usepackage{microtype}
    \usepackage{placeins}
    \usepackage{float}
    \usepackage{subcaption}
    \graphicspath{{dual_path_pool_prediction/}}

\begin{document}
    \let\WriteBookmarks\relax
    \def\floatpagepagefraction{0.7}
    \def\textpagefraction{.001}

    % Override page style: remove rule and text, keep only centered page number
    \makeatletter
    \def\ps@cas{%
      \def\@oddhead{}%
      \def\@evenhead{}%
      \def\@oddfoot{\hfil\small\thepage\hfil}%
      \def\@evenfoot{\hfil\small\thepage\hfil}%
    }%
    \def\ps@first{%
      \def\@oddhead{}%
      \def\@evenhead{}%
      \def\@oddfoot{\hfil\small\thepage\hfil}%
      \def\@evenfoot{\hfil\small\thepage\hfil}%
    }%
    \pagestyle{cas}%
    
    % Strip 50% opacity/tint from author group colors to keep name colors uniform
    \let\orgtextcolor\textcolor
    \def\textcolor{\@ifnextchar[\@textcolor@opt\@textcolor@noopt}
    \def\@textcolor@opt[#1]#2#3{\orgtextcolor[#1]{#2}{#3}}
    \def\@textcolor@noopt#1#2{%
      \expandafter\@check@tint#1!50\@nil{#1}{#2}%
    }
    \def\@check@tint#1!50#2\@nil#3#4{%
      \ifx\\#2\\%
        \orgtextcolor{#3}{#4}%
      \else
        \orgtextcolor{black}{#4}%
      \fi
    }
    \def\printorcid{}
    \makeatother
    
    \shorttitle{MamBOA: State-Space Architecture for Video Recognition}
    
    \shortauthors{M. B. \c{C}elik}
    
    \title [mode = title]{MamBOA: A State-Space Architecture for Differential Motion Synthesis in Video Recognition}  
    
    \author[1]{Mustafa Bora \c{C}elik}
    \cormark[1]
    \ead{mustafa.celik1@std.ankaramedipol.edu.tr}
    \credit{Conceptualization, Methodology, Software, Validation, Writing - Original draft preparation}
    
    \affiliation[1]{organization={Ankara Medipol University},
                    city={Ankara},
                    postcode={06050},
                    country={Turkey}}
    
    \cortext[1]{Corresponding author}
    
    \begin{abstract}
    Fine-grained action recognition demands temporal reasoning that general-purpose architectures address through different cost-accuracy tradeoffs: 3D dense operators couple computation to the input volume, while difference-based methods approximate motion through rigid, hand-crafted subtraction of uncontextualized features — each reflecting a deliberate design choice with corresponding limitations in expressiveness or flexibility. We present MamBOA, a backbone-agnostic temporal framework built upon a novel interleaved scan structure that recasts the selective state-space recurrence ($S6$) as a native motion synthesizer. By interleaving consecutive feature representations extracted from a pretrained backbone into a single alternating sequence, the proposed scan structurally drives the recurrence to encode both temporal observations of each position within a shared hidden state, separated by only a single decay step — rendering the inter-frame transition an intrinsic component of the state dynamics rather than an externally computed quantity. A cascade of dedicated alignment and decoding operations then distills this joint encoding into an explicit motion representation, which a dual-path pooling mechanism adaptively aggregates by balancing attention-driven selection with uniform temporal coverage. The framework interfaces seamlessly with CNN, Transformer, and Mamba backbone families, adding only $\sim$2.1 GFLOPs per feature pair. On Diving48, MamBOA achieves 85.02\% Top-1 accuracy with an image-pretrained backbone and 86.24\% with a video-pretrained backbone processing the entire video in a single forward pass — demonstrating that structurally induced state-space dynamics constitute a principled and general foundation for motion modeling.
    \end{abstract}
    
    
    \maketitle
    
    \section{Introduction}
    Video understanding requires not only recognizing what appears in individual spatial representations, but modeling how those representations evolve over time — a problem fundamentally rooted in sequential state dynamics. Appearance-based approaches such as TSN \cite{tsn} treat video as an average of independent segments, achieving strong results on coarse recognition tasks while falling short on actions whose discriminative signal lies in temporal dynamics rather than static content. This gap has driven two broad directions in the standalone end-to-end literature. On one side, heavy 3D convolutional networks (e.g., SlowFast \cite{slowfast}) and dense self-attention architectures (e.g., TimeSformer \cite{timesformer}, VideoSwin \cite{videoswin}) capture rich motion information but introduce a severe parameter footprint or a quadratic computational explosion when capturing long-form interactions. On the other side, lightweight difference methods such as TDN \cite{tdn} approximate motion through rigid explicit subtraction between adjacent feature maps — a fixed, non-learnable operation that cannot adapt to complex or multi-scale motion patterns, and whose discriminative capacity is bounded by the quality of the subtraction signal itself. State-space models \cite{mamba, vmamba} have recently emerged as a principled framework for efficient sequential modeling, offering continuous hidden state dynamics that can, in principle, natively accumulate and propagate temporal information — yet existing video applications \cite{videomamba, vis4mer} largely repurpose their recurrent dynamics for appearance aggregation, leaving the motion modeling potential of the state transition itself unexploited.

    We observe that this tension between parameter cost and temporal expressiveness shares a common root: motion is treated either as a quantity to be computed via heavy dense operators or as a fixed geometric subtraction, rather than a property that can be natively accumulated through the continuous hidden state dynamics of a sequential model. We introduce MamBOA, an architecture built on a different premise. By interleaving patches from two temporal feature maps into a single sequence before state-space processing, we allow the $S6$ recurrence to build an entangled joint encoding of both temporal positions. The motion differential is then natively accumulated and propagated through the evolution of the continuous hidden state dynamics ($h_t$) of the $S6$ recurrence, without requiring explicit geometric subtraction. Unlike traditional methods that subtract raw pixels or uncontextualized features, this mechanism operates on the deeply contextualized hidden states of the recurrence. Furthermore, by handling static spatial appearance in a parallel branch, the motion branch is structurally liberated to strictly capture temporal dynamics without information loss.
    
    Crucially, MamBOA is designed as a backbone-agnostic neural primitive that seamlessly integrates into existing architectures at the feature map level, enabling complex temporal reasoning without disrupting pre-trained spatial representations. MamBOA resolves temporal ambiguity by decomposing each feature map pair into a differential motion representation and a complementary spatial representation. The motion branch performs differential representation synthesis, mapping the interleaved spatial inputs into a distinct temporal subspace through the Interleaved Temporal Phase-shift Synthesis (ITPS) pipeline. Both representations are progressively aligned through bidirectional cross-conditioning within a modular processing stack, and temporal aggregation is performed by a dual-path pooling mechanism that balances global context with attention-weighted critical moments. Designed with efficiency in mind, MamBOA's differential synthesis operates with a marginal computational overhead of only $\sim$2.1 GFLOPs per feature pair, providing a lightweight temporal modeling head that scales efficiently with different backbones.
    
    We evaluate MamBOA on Diving48 \cite{diving48}, a strict fine-grained benchmark where categories are defined by combinations of rapid temporal attributes, requiring models to distinguish actions that are visually identical in static frames but dynamically distinct. To maintain a rigorous evaluation, our benchmark focuses exclusively on standalone end-to-end RGB networks, omitting auxiliary multi-modal inputs. Under standard evaluation protocols, MamBOA demonstrates highly competitive performance across disparate backbone families. The image-pretrained variant achieves 85.02\% Top-1 accuracy, while the video-pretrained variant delivers 86.24\% Top-1 accuracy over the full video sequence, bridging the gap between state-of-the-art representation capacity and low-overhead modular deployment.
    
    Our main contributions are summarized as follows:
    \begin{itemize}
    \item \textbf{SSM-Native Differential Representation Synthesis:} We introduce a novel state-space architecture that directly leverages interleaved sequential transitions to natively synthesize semantically meaningful differential motion representations from a pair of RGB feature maps — without optical flow, without rigid, hand-crafted frame subtraction, and without dense 3D operators. The core mechanism interleaves patches from two temporal feature maps into a single sequence, allowing the $S6$ recurrence to build an entangled joint temporal encoding. The motion differential is then natively accumulated and propagated through the evolution of the continuous hidden state $h_t$, and subsequently collapsed into a pure temporal representation via a learned projection. The synthesis process is described through four named functional stages (interleaved scan construction, phase-shift alignment, axial interaction, and motion differential computation) solely to facilitate precise exposition; together they constitute one unified pipeline serving this single objective.
    \item \textbf{Clip-Level Extension and Full-Video Coverage:} We show that the same differential representation synthesis principle applies at the clip level. Paired representations from a video-pretrained backbone — each encoding a rich multi-frame receptive field within a compact token — retain inter-clip differential structure that the proposed mechanism can resolve without modification. Combined with an adaptive-stride sampling strategy, this enables full-video temporal modeling whose computational cost scales with the number of clip pairs rather than with the total frame count.
    \item \textbf{Backbone-Agnostic Neural Primitive and Hardware-Aware Efficiency:} MamBOA serves as a backbone-agnostic neural primitive that can be directly applied to feature maps extracted from diverse architectural families, including CNN-, Transformer-, and Mamba-based backbones, enabling complex temporal reasoning without disrupting the pre-trained spatial representations. It produces consistent and meaningful differential representations across these heterogeneous representation spaces, establishing a unified temporal modeling principle. This architectural flexibility is achieved with a low computational cost of approximately $\sim$2.1 GFLOPs per feature-map pair, ensuring lightweight temporal modeling regardless of the underlying backbone footprint.
    \end{itemize}
    \begin{figure*}[!t]
    \centering
    \includegraphics[width=0.9\textwidth]{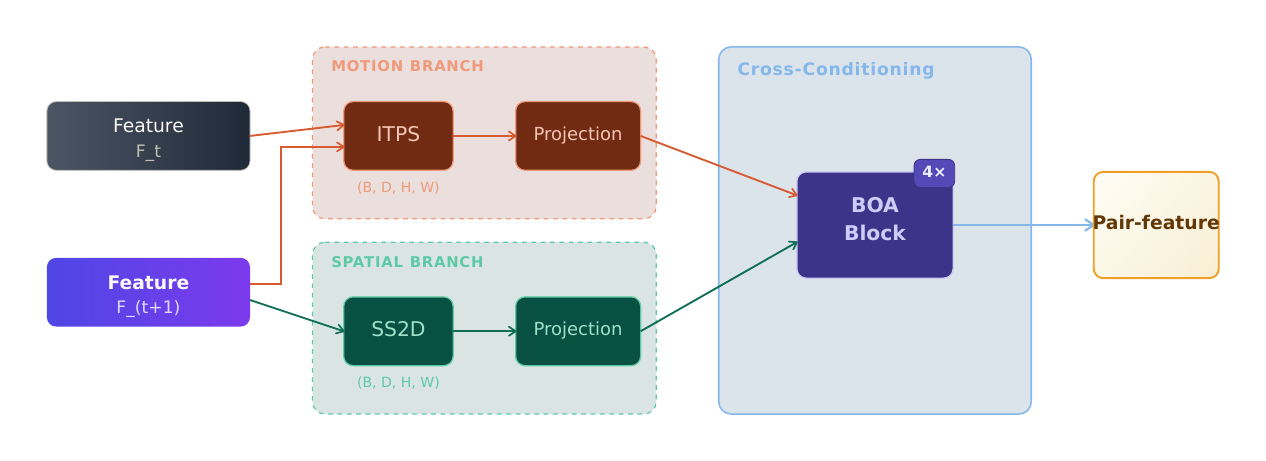}
    \caption{Overview of the proposed MamBOA architecture.}
    \label{fig:mamboa-overview-v2}
    \end{figure*}
    \section{Related Work}
    
    \paragraph{Temporal modeling in video recognition.}
    Early deep architectures for action recognition extended 2D image models
    along the temporal axis, either by aggregating frame-level predictions
    through sparse sampling, as in TSN \cite{tsn}, or by inflating spatial
    kernels into spatio-temporal 3D convolutions \cite{kinetics}. The latter
    family, exemplified by SlowFast \cite{slowfast}, captures rich motion
    dynamics but couples the computational cost directly to the
    spatio-temporal input volume. Transformer-based architectures such as
    TimeSformer \cite{timesformer}, ViViT \cite{vivit}, VideoSwin
    \cite{videoswin}, and MViTv2 \cite{mvitv2} replaced convolutional
    aggregation with dense self-attention, achieving strong accuracy at the
    price of quadratic complexity in the number of tokens. A complementary
    line of work pursues efficiency by augmenting 2D backbones with
    lightweight temporal operators: TSM \cite{tsm} shifts channels across
    time at zero FLOP cost, while GST \cite{gst} decomposes spatial and
    temporal convolutions into parallel groups. MamBOA follows the efficient
    end of this spectrum, but differs in deployment philosophy: rather than
    modifying the backbone internally, it operates as a modular temporal head
    on extracted feature map pairs, leaving the pre-trained spatial
    representation intact and keeping the temporal overhead fixed
    ($\sim$2.1 GFLOPs per pair) regardless of the backbone footprint.
    
    \paragraph{Motion representations and temporal differences.}
    Explicitly representing motion has a long history in action recognition.
    Two-stream networks \cite{twostream} supply optical flow as a separate
    input modality, at the cost of expensive flow computation and a doubled
    inference pipeline. To avoid this, difference-based methods approximate
    motion internally: TDN \cite{tdn} computes explicit subtractions between
    adjacent feature maps at multiple scales, and TEA \cite{tea} uses
    feature-level differences to excite motion-sensitive channels. While
    effective, these mechanisms share a structural limitation: the
    differential operator itself is a fixed, hand-crafted subtraction applied
    to uncontextualized features, so the quality of the motion signal is
    bounded by what a rigid pointwise difference can express. MamBOA departs
    from this paradigm at the representational level. The two temporal
    observations are first entangled within the hidden state of a selective
    recurrence, which contextualizes each position with accumulated spatial
    history and input-dependent gating; the differential is then decoded from
    this joint encoding by a learnable asymmetric projection. The subtraction
    in MamBOA therefore acts on deeply contextualized state trajectories
    rather than raw features, and its scale coefficients are learned rather
    than fixed.
    
    \paragraph{State-space models in vision and video.}
    Structured state-space models were introduced for long-sequence modeling
    by S4 \cite{s4}, and Mamba \cite{mamba} made the recurrence
    input-selective, achieving linear-time sequence modeling with
    content-dependent state transitions. In the visual domain, Vim \cite{vim}
    and VMamba \cite{vmamba} adapted the selective scan to images through
    bidirectional and multi-directional traversals of the patch grid. For
    video, ViS4mer \cite{vis4mer} applies state-space layers to aggregate
    long-form clip features, and VideoMamba \cite{videomamba} extends the
    bidirectional scan over the joint spatio-temporal token volume. A common
    property of these video applications is that the recurrence is employed
    as an efficient \emph{aggregation} operator: tokens from all frames are
    merged into a unified representation in which appearance and motion
    remain entangled. In contrast, MamBOA exploits a complementary and, to
    the best of our knowledge, previously unexplored property of the
    selective recurrence: by structurally arranging the input as an
    interleaved sequence of temporally paired patches, the hidden state is
    forced to encode the local transition between two observations of the
    same spatial location, turning the scan itself into a differential motion
    synthesizer rather than an appearance aggregator.
    
    \paragraph{Adapting image models and fine-grained recognition.}
    A recent direction reuses strong image-pretrained models for video with
    minimal modification: SIFAR \cite{sifar} rearranges sampled frames into a
    super-image consumed by an unmodified image classifier, and dual-path
    adaptation methods \cite{dualpath} attach lightweight temporal adapters
    to frozen image transformers. While these hybrid and adaptation-based pipelines
    confirm that pre-trained spatial representations carry substantial value for video,
    they often rely on specific input formatting (e.g., heuristic sampling or image rearrangement)
    or restrict temporal reasoning strictly to adapter capacity. This functionally distinguishes them 
    from foundational end-to-end video architectures. Fine-grained benchmarks expose 
    this limitation most clearly: in Diving48 \cite{diving48},
    categories share nearly identical static appearance and differ only in
    the composition of rapid temporal attributes, so scene and object cues
    provide no shortcut \cite{diving48, ssv2}. MamBOA targets precisely this
    regime. It preserves the pre-trained spatial representation as a frozen
    semantic substrate, while a dedicated state-space differential mechanism
    --- operating as a unified temporal head rather than a localized adapter --- supplies the temporal
    discrimination that adaptation-based methods delegate to limited modules.
    
    \section{Methodology}
    
    \subsection{Overview}
    
    We propose MamBOA, a motion-aware video representation architecture grounded in the sequential state dynamics of selective state-space models. The model operates on pairs of feature maps extracted from consecutive temporal samples and decomposes each pair into two complementary representations: a differential motion representation that performs temporal subspace projection, and a spatial representation that preserves static appearance. Given a pair of samples, the features are extracted from a shared backbone and routed into two parallel branches. The core of our approach lies in the conceptual division of labor between these branches. The motion branch first entangles the two temporal positions via an interleaved scan, allowing the $S6$ recurrence to accumulate a joint temporal encoding through its hidden state evolution. It then isolates the pure motion signal by applying a learned projection that cancels out the shared static background. Crucially, because the parallel spatial branch is dedicated exclusively to modeling spatial appearance, the motion branch is structurally liberated from preserving static context, allowing it to fully discard non-dynamic information. The two branches are progressively aligned through bidirectional cross-conditioning within BOA blocks, fused into compact pair-level features, and aggregated by a Dual-Path temporal pooling module into a single video-level prediction. An overview of the architecture is illustrated in Fig.~\ref{fig:mamboa-overview-v2}.
    
    \begin{figure*}[!t]
    \centering
    \includegraphics[width=0.85\textwidth]{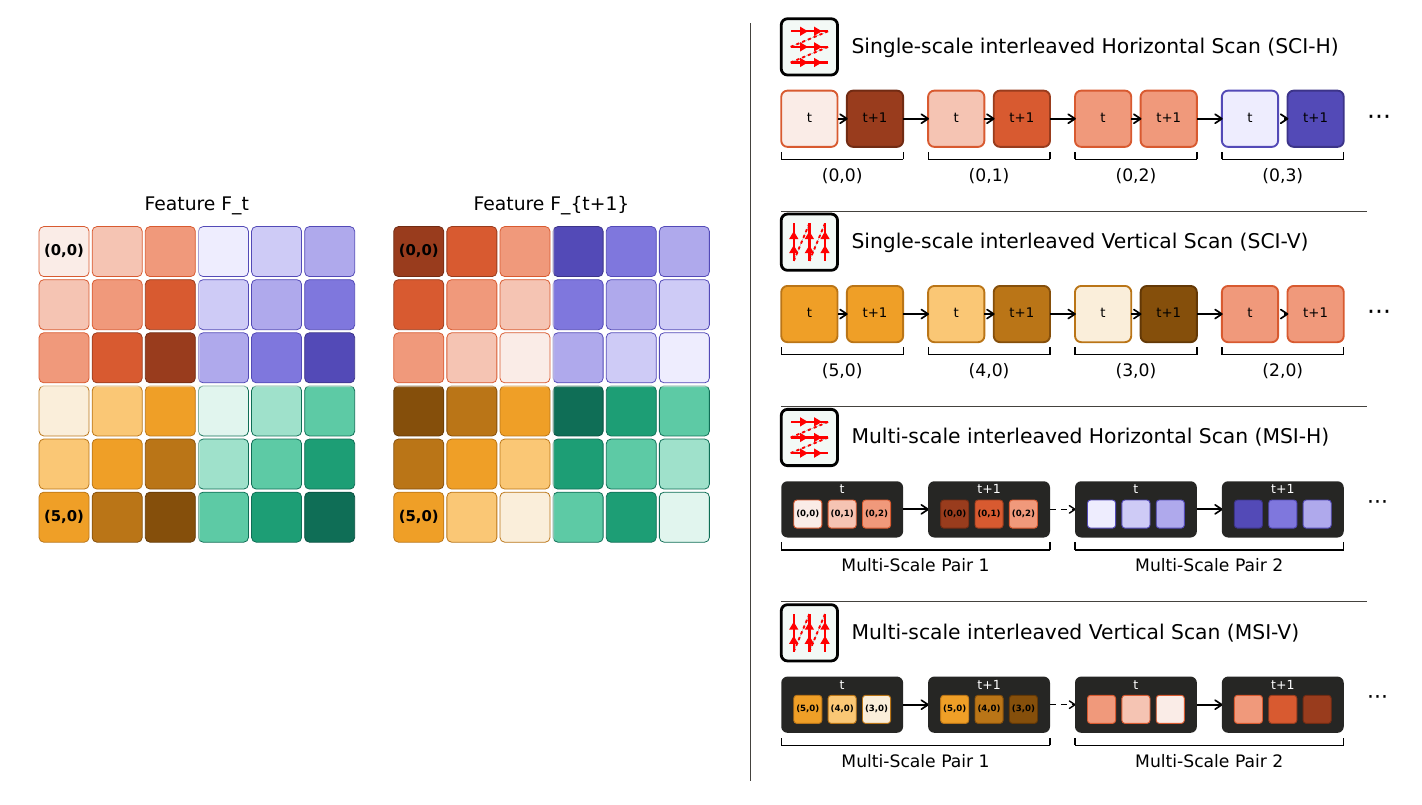}
    \caption{Illustration of the single-scale and multi-scale interleaved scan patterns, along with the canonical row-major reordering applied prior to ITPS processing.}
    \label{fig:interleaved-scan}
    \end{figure*}

    \subsection{Interleaved Scan and Feature Pair Construction}
    The model receives pairs of feature maps $\mathbf{F}^{(0)}, \mathbf{F}^{(1)} \in \mathbb{R}^{C \times H \times W}$ extracted from a shared backbone at two temporal positions. For the image-pretrained variant, these correspond to sparsely sampled frames; for the clip-level variant, they correspond to consecutive clip representations. In both cases, the model operates uniformly on these feature map pairs, providing a consistent mechanism for extracting differential motion regardless of whether the inputs are discrete frames or clip-level summaries.
    To prepare these feature maps for state-space processing, we flatten the spatial grid into a sequence of length $L$ ($H \times W = L$), resulting in input tensors of shape $\mathbb{R}^{B \times L \times C}$. We then apply an interleaved scan. Patches from the two temporal positions are interleaved into a single alternating sequence $\{x_1, x_2, \dots, x_{2L}\}$, forming a joint tensor of shape $\mathbb{R}^{B \times 2L \times C}$, where odd and even positions alternate strictly between $\mathbf{F}^{(0)}$ and $\mathbf{F}^{(1)}$. The S6 recurrence processes this sequence as:
    \begin{equation}
    h_k = \bar{A}_k h_{k-1} + \bar{B}_k x_k, \qquad y_k = C_k h_k
    \label{eq:s6_state}
    \end{equation}
    We employ two structural variations of this interleaving: a single-scale scan and a multi-scale scan. Both the single-scale and multi-scale structures are scanned in horizontal (row-major) and vertical (column-major) directions, yielding a total of four distinct scan outputs.
    Reverse scans are deliberately omitted to enforce a strict temporal arrow in the motion representation. While standard spatial vision models rely on bidirectional scans to gather symmetric context, motion is inherently causal. The forward interleaved scan explicitly aligns with this causality by ordering the past observation ($t-1$) immediately before the present ($t$). Computing reverse scans would simultaneously encode the non-causal backward transition ($t \to t-1$). By omitting them, we constrain the network to learn exclusively forward-time dynamics, a structural prior that concurrently halves the computational cost of the scan process. The scan patterns are illustrated in Fig.~\ref{fig:interleaved-scan}.
    \paragraph{Single-scale scan.}
    In the single-scale variant, consecutive tokens in the sequence correspond to the same spatial location observed at the two temporal positions. Expanding the recurrence at an even step $k=2j$:
    \begin{equation}
    h_{2j} = \bar{A}_{2j}\bar{A}_{2j-1}h_{2j-2} + \bar{A}_{2j}\bar{B}_{2j-1}x_{2j-1}^{(0)} + \bar{B}_{2j}x_{2j}^{(1)}
    \label{eq:single_scan}
    \end{equation}
    The critical structural property of the interleaved arrangement is that the two temporal observations at the same spatial location $j$ — defined in the sequence as $x^{(0)}_j \equiv x_{2j-1}^{(0)}$ and $x^{(1)}_j \equiv x_{2j}^{(1)}$ — are always adjacent, separated by exactly one recurrence step. Consequently, $x^{(0)}_j$ contributes to $h_{2j}$ via the term $\bar{A}_{2j}\bar{B}_{2j-1}x^{(0)}_j$, attenuated by only a single decay factor $\bar{A}_{2j}$. In a standard sequential (position-first) scan, by contrast, the same spatial location of temporal position $t{-}1$ would be $L$ steps away from its counterpart at temporal position $t$, reducing its cross-position contribution by $\bar{A}^L \to 0$ for any non-trivial sequence length $L$. The interleaved arrangement therefore structurally guarantees that $h_{2j}$ is a joint encoding of both temporal observations in direct succession: it retains the accumulated spatial context in $h_{2j-2}$ alongside the minimally-attenuated, input-gated contributions of both temporal positions. This entangled joint representation is what makes $h_{2j}$ structurally well-suited for the downstream motion decoding performed by MDM, regardless of whether the inputs are individual frame features or compact clip-level representations. We deliberately omit the native skip connection present in the standard S6 formulation. Reintroducing this residual connection would add a raw single-position patch token directly back to this joint encoding, systematically biasing $y_k$ toward per-position appearance at every recurrence step. Since the parallel SS2D branch is dedicated to modeling spatial appearance, the skip term is structurally incompatible with the joint temporal encoding objective.

    \begin{figure*}[!t]
    \centering
    \includegraphics[width=0.85\textwidth]{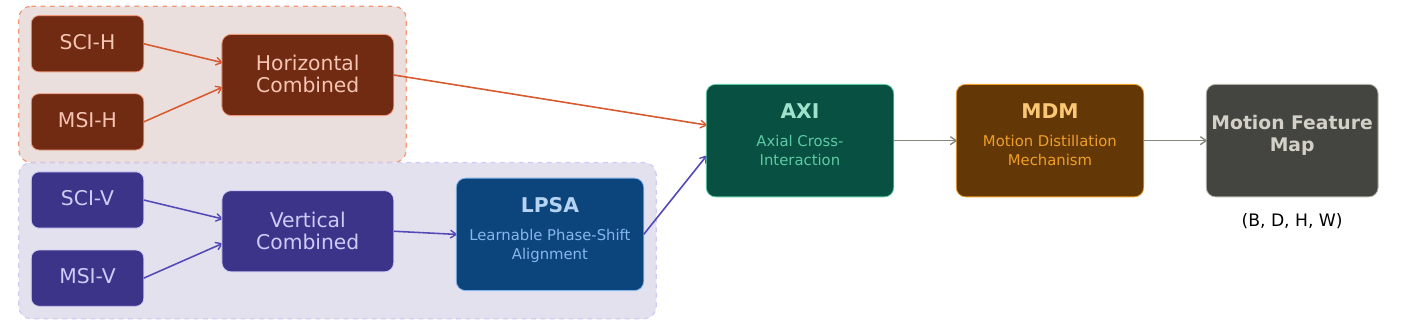}
    \caption{Overview of the Interleaved Temporal Phase-shift Synthesis (ITPS) pipeline.}
    \label{fig:fourier-shift-pipeline-v3}
    \end{figure*}

    \paragraph{Multi-scale scan.}
    The multi-scale variant groups $g$ neighboring patches from one temporal position before alternating with the corresponding group from the other ($g=3$ in our implementation). Within each $2g$-token block, the first $g$ tokens originate from $\mathbf{F}^{(0)}$ and the next $g$ from $\mathbf{F}^{(1)}$. The hidden state accumulated after processing the first $g$ tokens from temporal position $t{-}1$ constitutes a \emph{pre-transition} spatial representation:
    \begin{equation}
    h_{\text{pre}} = \sum_{r=0}^{g-1}\left(\prod_{m=r+1}^{g-1}\bar{A}_m\right)\bar{B}_r\,x^{(0)}_r
    \label{eq:pre_transition}
    \end{equation}
    where the product is empty (equal to 1) when $r=g-1$. After the full $2g$-token block, the hidden state takes the form:
    \begin{equation}
    h_{\text{post}} = \underbrace{\left(\prod_{m=g}^{2g-1}\bar{A}_m\right)h_{\text{pre}}}_{\text{decayed spatial prior}} + \underbrace{\sum_{s=0}^{g-1}\left(\prod_{m=g+s+1}^{2g-1}\bar{A}_m\right)\bar{B}_{g+s}\,x^{(1)}_s}_{\text{temporal transition terms}}
    \label{eq:multi_scan}
    \end{equation}
    The first term in Eq.~\ref{eq:multi_scan} is the pre-transition spatial state $h_{\text{pre}}$ decayed by an additional $g$ recurrence steps. As $g$ grows, $h_{\text{pre}}$ itself integrates a progressively wider spatial neighborhood (Eq.~\ref{eq:pre_transition}) before the cross-position transition occurs. Because the state accumulates more broad spatial context prior to the temporal switch, the representation naturally shifts from being highly temporally-sensitive toward becoming spatially-enriched, while the shared spatial background (the pool, the platform) increasingly cancels in the downstream MDM step.

    This analysis establishes two boundary conditions on $g$. At the lower bound, $g=2$ produces a pre-transition state that integrates only two immediately adjacent patches — a spatial footprint so narrow that its accumulated context is nearly indistinguishable from the single-patch accumulation of the single-scale scan, rendering the two scales informationally redundant. At the upper bound, as $g$ approaches the spatial extent of the feature map, each group approximates a global spatial summary of an entire temporal position before any temporal interaction occurs, and the resulting representation is dominated by static, position-level appearance.

    The motivation for simultaneously using single-scale and multi-scale scans is to capture two complementary aspects of video motion dynamics: the single-scale scan targets localized, instantaneous point-wise transitions, while the multi-scale scan exposes structured motion patterns that unfold across a spatial neighborhood — such as regional body trajectories or coherent limb movements spanning multiple patches. This complementarity requires the two representations to be structurally distinct. Group size $g=3$ is the minimal configuration that satisfies this requirement: it is wide enough to diverge meaningfully from single-scale behavior and expose genuine neighborhood-level temporal structure, yet narrow enough to preserve the temporal sensitivity of the hidden state at the transition boundary. We accordingly adopt $g=3$ as the multi-scale group size.
    Before entering the ITPS pipeline, each scan sequence is mapped back to its canonical $(H,W)$ spatial coordinates through an inverse traversal. This operation not only ensures that tokens from horizontal and vertical scans at the same spatial position are aligned, but importantly, maps the outputs of both single- and multi-scale scans back into a strict (position, time) alternating order, cleanly interleaving the two temporal positions at every spatial location.
    
    \subsection{ITPS: Interleaved Temporal Phase-shift Synthesis}
    \label{sec:itps}

    Motion in video is a vector quantity: it has both magnitude and direction. The four scan sequences produced in the previous stage capture horizontal and vertical motion components separately, each from two scale perspectives. These are not two independent signals — they are orthogonal projections of the same underlying displacement field. Just as knowing the $x$- and $y$-projections of a displacement vector independently does not reveal the true motion direction without bringing them into a common reference frame, the horizontal and vertical scan outputs must be spatially aligned before their joint interaction can expose multi-directional motion structure. The ITPS pipeline performs this alignment and synthesizes a single differential motion feature map through sequential stages, as illustrated in Fig.~\ref{fig:fourier-shift-pipeline-v3}.
    \paragraph{Scale-axis combination.}
    The single-scale and multi-scale scan outputs along each axis are first merged through learnable coefficients, producing a single horizontal and a single vertical signal:
    \begin{equation}
    S_H = \gamma_H S_{\text{SCI-H}} + (1-\gamma_H) S_{\text{MSI-H}}
    \label{eq:scale_combine_h}
    \end{equation}
    \begin{equation}
    S_V = \gamma_V S_{\text{SCI-V}} + (1-\gamma_V) S_{\text{MSI-V}}
    \label{eq:scale_combine_v}
    \end{equation}
    where $\gamma_H, \gamma_V \in [0,1]$ are learned per-channel. At this point, $S_H$ and $S_V$ contain the horizontal and vertical temporal features. To prepare them for 1D phase shifting, the 2D feature maps $S_H$ and $S_V$ are flattened into 1D row-major sequences. However, perfectly matching their 1D indices is not enough: because $S_H$ was accumulated via a horizontal scan and $S_V$ via a vertical scan, the causal 1D S6 filters have inherently skewed their spatial receptive fields in orthogonal directions. Consequently, at any given token index, $S_H$ and $S_V$ are functionally centered on slightly different spatial coordinates. This systematic receptive field misalignment must be corrected before the two axes can be meaningfully combined.
    \paragraph{LPSA: Learnable Phase-Shift Alignment.}
    The exact spatial offset between the receptive fields of $S_H$ and $S_V$ depends on the scan geometry and is generally a non-integer, channel-dependent quantity. We perform this alignment in the frequency domain. The Fourier shift theorem states that a spatial translation of $\tau$ positions in a 1D sequence of length $L_{\text{seq}}$ corresponds to a phase rotation $\exp\left(-i2\pi f\tau/L_{\text{seq}}\right)$ at frequency index $f$. We apply a real 1D FFT along the sequence dimension ($L_{\text{seq}}=2L$, representing the full interleaved token sequence for the two temporal positions). This approach has three critical advantages. First, the learned parameter $\tau$ is continuous, providing sub-pixel spatial translation precision inaccessible to integer shifts. Second, it is global, operating on the entire spectrum simultaneously. Third, we learn $\tau$ independently per channel, allowing different feature channels to discover their optimal spatial alignment offsets. 

    We hold the horizontal signal $S_H$ fixed as the reference anchor and shift only the vertical signal $S_V$. For each frequency index $f$, the channel-wise parameter $\tau$ defines the phase angle:
    \begin{equation}
    \theta(f) = -\frac{2\pi f\,\tau}{L_{\text{seq}}}
    \label{eq:lpsa_theta}
    \end{equation}
    The spectrum is rotated and inverted back to the 1D token domain:
    \begin{equation}
    \hat{S}_V = \mathcal{F}^{-1}\left(\mathcal{F}(S_V)(f)\cdot \exp\left(i\theta(f)\right)\right)
    \label{eq:lpsa_shift}
    \end{equation}
    While the 1D Fourier shift naturally imposes a circular boundary condition (wrap-around) on the sequence, the learned translations $\tau$ converge to highly fractional, sub-pixel values (empirical measurements show a mean absolute magnitude of $|\tau| \approx 0.35$). Consequently, any circular wrap-around artifacts are strictly confined to the extreme sequence boundaries, while the vast majority of the tokens benefit from continuous, artifact-free spatial interpolation. Furthermore, while a 1D shift on a row-major sequence primarily constitutes a horizontal spatial translation, this learnable shift successfully corrects the dominant component of the misalignment. After this step, the receptive fields of $S_H$ and $\hat{S}_V$ are spatially aligned.
    \paragraph{AXI: Axial Interaction.}
    With the receptive fields of the two components aligned, their element-wise product exposes regions of complex, diagonal co-activation. A motion event that activates both $S_H$ and $\hat{S}_V$ simultaneously produces a strong interaction signal, whereas isolated horizontal or vertical motion does not. Rather than explicitly computing geometric angles (which would require operations like $\arctan2$), we employ the aligned vertical signal as a spatial gating mechanism. A learnable coefficient $\alpha$ blends this co-activation signal with the primary horizontal signal:
    \begin{equation}
    S_{\text{AXI}} = \alpha\,(S_H \odot \hat{S}_V) + (1-\alpha)\,S_H
    \label{eq:axi}
    \end{equation}
    The parameter $\alpha$ controls the strength of this gating. Through this interaction, the network learns to dynamically emphasize regions of multi-directional motion while preserving the foundational horizontal motion representation.
    \paragraph{MDM: Motion Decoding Module.}
    At this stage, $S_{\text{AXI}}$ (still possessing the $\mathbb{R}^{B \times 2L \times C}$ sequence shape) jointly encodes both temporal positions in an entangled state. To collapse this sequence into a single 2D feature map for the subsequent BOA stack, we explicitly de-interleave the alternating sequence to isolate the two temporal positions. Specifically, the sequence is reshaped into $\mathbb{R}^{B \times L \times 2 \times C}$, where the inner dimension cleanly separates the alternating tokens into past and present. These separated tokens are then projected into a unified tensor via a learnable decoding step:
    \begin{equation}
    M = \omega_1\,Z_1 - \omega_0\,Z_0
    \label{eq:mdm}
    \end{equation}
    where $Z_0$ and $Z_1$ are the perfectly de-interleaved token segments of $S_{\text{AXI}}$ corresponding to the reference and present feature maps, respectively. It is critical to note that the operation in Equation~\ref{eq:mdm} should not be confused with the rigid, hand-crafted geometric subtractions used in prior difference networks. The actual temporal transitions have already been inherently encoded by the hidden states of the S6 recurrence during the earlier interleaved scan (which generates the entangled states $Z_0$ and $Z_1$). Here, the learnable projection serves as the primary motion extraction stage: because both $Z_0$ and $Z_1$ share the same accumulated spatial context, their subtraction effectively cancels out static spatial interference (e.g., unchanged background or static objects), isolating the pure temporal change. While one might consider standard downsampling methods (such as mean or max pooling) to collapse the sequence, they are structurally inadequate: motion is fundamentally an asymmetric, directional quantity. Symmetric pooling would blur $Z_0$ and $Z_1$ into a static spatial summary, obliterating the temporal directionality. By utilizing a learnable, asymmetric projection, MDM successfully cancels the static spatial context and collapses the sequence into a 2D motion feature map $\mathbf{M} \in \mathbb{R}^{C \times H \times W}$, fully preserving the signed differential dynamics. This 2D spatial representation is then passed to the subsequent BOA stack.
    
    \subsection{Spatial Branch}
    In parallel, the spatial branch explicitly isolates the tokens belonging to the present temporal position from the interleaved sequences. These present-position tokens are restored to a 2D grid and processed by a standard S6 state-space scan in four directions (SS2D), whose outputs are averaged to produce a clean spatial feature map. By extracting only the tokens of the present feature map, the spatial branch specializes in modeling the rich 2D structural appearance. Since our core architectural contribution lies in differential motion extraction, we employ a standard multi-directional S6 configuration for this spatial branch without further modification, acting as a straightforward but effective complement to the motion branch.
    \subsection{BOA Block}
    
    \begin{figure*}[!t]
    \centering
    \begin{subfigure}[b]{0.38\textwidth}
      \centering
      \includegraphics[trim=0 0 180 0, clip, width=\textwidth]{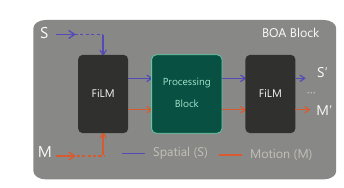}
      \caption{BOA block architecture.}
      \label{fig:bora-block-detail}
    \end{subfigure}
    \hspace{1.2cm}
    \begin{subfigure}[b]{0.43\textwidth}
      \centering
          \includegraphics[trim=173 0 0 0, clip, width=\textwidth]{concat.pdf}
      \caption{Dual-Path Temporal Pooling mechanism.}
      \label{fig:dual-path-pool}
    \end{subfigure}
    \caption{Detailed schematic of the MamBOA architecture components.}\label{fig:mamboa-components}
    \end{figure*}
    
    Before deep fusion, both branches are projected into a shared latent space. Within each BOA block, the spatial features are modulated by parameters (scale and shift) derived from the motion features, and the motion features are likewise conditioned by the spatial features. This bidirectional FiLM \cite{film} conditioning lets the two representations gradually recognize and align with one another. Formally, this bidirectional modulation is defined as:
    \begin{gather}
    S' = \gamma_s \cdot \text{Norm}(S) + \beta_s, \quad \gamma_s, \beta_s = \text{Linear}(M) \label{eq:film_s} \\
    M' = \gamma_m \cdot \text{Norm}(M) + \beta_m, \quad \gamma_m, \beta_m = \text{Linear}(S) \label{eq:film_m}
    \end{gather}

    where $\text{Norm}$ denotes Layer Normalization, and $\text{Linear}$ represents linear projections. Following modulation, the features $S'$ and $M'$ are not immediately concatenated; rather, they are independently passed through a modern processing block to refine their respective representations. A defining characteristic of BOA module is its processing-block-agnostic design; its formulation does not rely on paradigm-specific operations. While the framework is designed to seamlessly integrate with various block topologies, we adopt the ConvNeXt-V2 block \cite{convnextv2} as the primary instantiation in our architecture due to its strong empirical performance. As demonstrated in our ablation studies, this flexibility allows the BOA module to operate effectively with modern blocks from diverse architectural families without requiring structural modifications. This sequence of bidirectional conditioning followed by parallel block processing constitutes a single BOA layer. The detailed schematic of the BOA block architecture is shown in Fig.~\ref{fig:bora-block-detail}. The entire stack is repeated four times, allowing progressively deeper and more intricate interactions between the motion and spatial representations before they are finally concatenated in the deep fusion stage.

    \paragraph{Feature Fusion.}
    The final spatial and motion features are concatenated along the channel dimension and fused by a $1\times 1$ convolution with normalization and activation, preserving the 2D spatial structure. This joint representation encodes the local transitions of the temporal feature-map pair at every spatial location, and is subsequently passed to the spatial-aware pooling and classification stages.

    \subsection{Dual-Path Temporal Pooling}
    
    Repeating the above for every sampled pair yields a temporal sequence of feature vectors  $\{\mathbf{F}_t\}_{t=1}^T$. Reducing this sequence to a single summary is non-trivial: pure mean pooling drowns the few critical moments in the surrounding static content, while pure attention pooling tends to collapse onto one or two pairs and discard global context. We therefore adopt a dual-path scheme, as illustrated in Fig.~\ref{fig:dual-path-pool}.
    
    The attention path computes per-timestep scores $s_t$, normalizes them with a softmax function to obtain attention weights $\alpha_t$, and produces a weighted sum $f_{\text{attn}}$ that isolates the most decisive moments:
    \begin{equation}
    s_t = \text{Linear}(\mathbf{F}_t), \quad \alpha_t = \frac{\exp(s_t)}{\sum_i \exp(s_i)}
    \label{eq:attn_weights}
    \end{equation}
    \begin{equation}
    f_{\text{attn}} = \sum_{t=1}^T \alpha_t \mathbf{F}_t
    \label{eq:f_attn}
    \end{equation}
    In parallel, the mean path averages all timesteps, summarizing temporally invariant structure such as athlete appearance and scene:
    \begin{equation}
    f_{\text{mean}} = \frac{1}{T}\sum_{t=1}^T \mathbf{F}_t
    \label{eq:f_mean}
    \end{equation}
    The two paths are blended through:
    \begin{equation}
    f_{\text{final}} = f_{\text{attn}} + 2\sigma(\beta) \cdot f_{\text{mean}}
    \label{eq:dual_path_blend}
    \end{equation}
    where $\beta$ is an unbounded learnable scalar, and $\sigma(\cdot)$ is the sigmoid function, making the effective weighting coefficient $2\sigma(\beta)$ bounded in $(0, 2)$. The network thus learns the trade-off between static and dynamic context: when the discriminative signal is highly concentrated in a few specific temporal moments, the optimization drives $\beta \to -\infty$, minimizing the mean path contribution so that attention dominates. Conversely, when global background context is informative, $\beta$ increases to draw heavily on the mean path. 
    
    To prevent the attention weights from collapsing onto a single frame, we apply an entropy regularization term on the attention weights during training.

    \begin{figure*}[!t]
    \centering
    \includegraphics[width=0.8\textwidth]{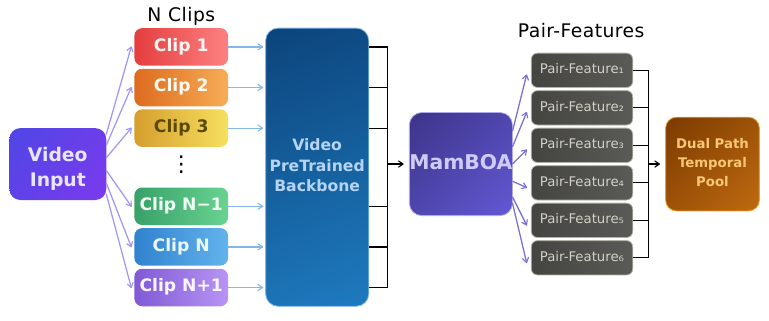}
    \caption{Clip-level extension of the MamBOA architecture.}
    \label{fig:videomamboa}
    \end{figure*}

    \paragraph{Cosine Classifier.}
    The pooled feature is classified by a cosine classifier, which $L_2$-normalizes both the feature and the weight matrix before producing logits. As is standard practice in fine-grained visual recognition literature \citep{sifar}, this normalization stabilizes the magnitude of the logits and significantly improves separation between classes by reducing intra-class variance.

    \subsection{Adaptive Clip-Level Temporal Coverage}
    
    Rather than operating on a fixed set of sparsely sampled frames, the video-pretrained variant extends the proposed differential framework to the clip level, as illustrated in Fig.~\ref{fig:videomamboa}. Given a video, $N$ temporal clips are extracted using an adaptive-stride sampling strategy and processed by a shared backbone. The resulting clip representations are then treated analogously to feature-map pairs, allowing the same MamBOA differential mechanism and temporal aggregation pipeline to operate at the clip level.
    
    Conventional video recognition architectures typically process a fixed number of frames (e.g., 8 or 16), causing computational cost to scale directly with the number of sampled frames while leaving a large portion of long videos unobserved. In contrast, the proposed formulation decouples temporal coverage from the backbone input length. Instead of selecting a small subset of frames, clips are distributed across the entire video duration, enabling substantially broader temporal coverage.
    
    For a video containing $T$ frames, clip starting positions are computed dynamically according to the video length. Let $L_c$ denote the temporal receptive field of a clip. The maximum valid starting position is
    \begin{equation}
    s_{\max}=T-L_c
    \label{eq:s_max}
    \end{equation}
    and the interval between adjacent clips is defined as
    \begin{equation}
    \Delta s=\frac{s_{\max}}{N-1}.
    \label{eq:delta_s}
    \end{equation}
    Consequently, clip locations automatically expand or contract according to the video duration, ensuring approximately uniform temporal coverage regardless of video length. During training, temporal jitter is applied around each nominal clip position for regularization, whereas deterministic positions are used during inference.
    
    Each clip contributes a prediction token that participates in the Dual-Path Temporal Pooling module, where temporal evidence from different parts of the video is aggregated into a unified representation. As a result, the computational cost of the temporal aggregation stage scales primarily with the number of clip-level prediction tokens rather than the total number of frames in the original video. This enables broad temporal coverage while maintaining a single-pass inference pipeline.
    
    To illustrate the coverage gain concretely: while conventional approaches observe only a fixed, small subset of frames, the proposed method distributes $N$ clips across the full temporal extent, each contributing a multi-frame receptive field. The resulting $N-1$ consecutive clip pairs thus aggregate temporal evidence from substantially more of the video than any single fixed-window sampling strategy could provide.
    
    As shown in the ablation study, increasing the number of clip-level Pair-Features consistently improves recognition accuracy, suggesting that richer temporal evidence accumulation directly benefits action understanding.
    
    \section{Experiments}
    
    \subsection{Dataset and Evaluation Protocol}
    We evaluate MamBOA on Diving48 \cite{diving48}, comprising 16,997 video clips across 48 categories defined by complex temporal attributes. While prominent datasets like Kinetics \cite{kinetics} and Something-Something V2 \cite{ssv2} are excellent for general representation learning, benchmark selection here is strictly driven by our core objective: isolating pure motion extraction. Since modern backbones already excel at extracting static cues, we require a testbed that nullifies spatial shortcuts. Diving48 shares a uniform visual environment across all classes, forcing the model to rely exclusively on long-range temporal dynamics rather than scene bias \cite{diving48}. This makes it the ideal environment to validate MamBOA's differential capabilities. We report Top-1 accuracy on the official test split (15,027 train / 1,970 test).
    
    This single-dataset focus is a deliberate experimental design decision. Because all 48 categories share an identical background (a diving pool), spatial shortcuts are structurally eliminated by the dataset's construction. As a consequence, every measured performance gain can be causally attributed to temporal modeling quality, yielding a controlled environment for deep mechanism validation. Evaluating generalization across broader dataset distributions is a natural complement to this mechanistic analysis and is left for future work.
    
    \subsection{Implementation Details}
    For both variants, features are extracted from intermediate layers of the backbones to maintain an optimal balance between high-level semantic context and the spatial resolution necessary for fine-grained motion modeling.
    \paragraph{Image-pretrained variant.} The backbone is VMamba-Base \citep{vmamba} with depths [2, 2, 4], pretrained on ImageNet-1K. Features are extracted from the output of the third stage. Input frames are resized to 224$\times$224 with standard augmentation (random resized crop, horizontal flip, color jitter) during training and center-cropped during inference. We train for 50 epochs using AdamW \cite{adamw} (backbone learning rate of 1e-5, head learning rate of 1e-4), cosine decay, and batch size 16. Online Hard Example Mining (OHEM) \citep{ohem} is applied from epoch 18 with keep ratio 0.7. MixUp augmentation is used with $\alpha=0.2$. The BOA stack consists of 4 ConvNeXt-V2 processing blocks. Inference uses 30 views (10 temporal $\times$ 3 spatial crops).
    
    \paragraph{Video-pretrained variant.} The backbone is MViT-V2-S \citep{mvitv2} pretrained on Kinetics-400, with features hooked at block 13. During training, we extract 7 temporal clips ($N=7$), each with 16 frames and a stride of 2. The model is trained for 45 epochs with a batch size of 32 using AdamW \cite{adamw} (backbone learning rate of 1e-5, head learning rate of 1e-4), utilizing a SequentialLR scheduler with a 5-epoch warmup and cosine decay. The backbone is frozen for the first 5 epochs. Augmentation and regularization follow the image variant, with label smoothing set to 0.1. OHEM is applied starting from epoch 18 (with a warmup of 5 epochs and a keep ratio of 0.7). A Spatial-Aware Temporal Attention Pool (reducing the temporal dimension $T=8 \to 1$, where the 16-frame input with stride 2 is inherently reduced to $T=8$ by MViT's temporal downsampling, via a zeros-initialized 3D Conv with kernel size 1, no bias, and softmax weighting) is used to project backbone features. No test-time view ensembling is applied; predictions are aggregated in a single forward pass.
    
    \paragraph{Attribute-specific pooling and hierarchical loss.}
Diving48 provides compositional action labels decomposed into four 
attribute dimensions: takeoff style, somersault count, twist count, and 
entry position. To exploit this structure, we augment the shared dual-path 
pool with four additional attribute-specific attention heads, each producing 
an independent pooled representation from the same logit sequence. A 
lightweight linear classifier is attached to each attribute head (takeoff: \text{tk}, somersault: \text{som}, twist: \text{tw}, entry position: \text{pos}) and 
supervised by the corresponding attribute label during training.
The combined training objective is:
\begin{equation}
    \mathcal{L} = \mathcal{L}_{\text{main}} 
    + 0.2\,\mathcal{L}_{\text{tk}} 
    + 0.3\,\mathcal{L}_{\text{som}} 
    + 0.3\,\mathcal{L}_{\text{tw}} 
    + 0.2\,\mathcal{L}_{\text{pos}}
    + \lambda_{\text{ent}}\,\mathcal{L}_{\text{ent}}
    \label{eq:hierarchical_loss}
\end{equation}
The auxiliary weights sum exactly to 1.0 ($0.2 + 0.3 + 0.3 + 0.2 = 1.0$). This constitutes a deliberate mathematical scale normalization (convex combination) that prevents the hierarchical loss from artificially inflating the total gradient magnitude, ensuring stable optimization without requiring global learning rate adjustments. The higher weights for somersault and twist reflect their more complex temporal dynamics compared to takeoff and entry position. Here, $\mathcal{L}_{\text{ent}} = -\sum_t \alpha_t\log\alpha_t$ is the entropy regularization term applied to the temporal attention weights to prevent collapse onto a single frame, and $\lambda_{\text{ent}} = 0.1$ is its scaling hyperparameter. All attribute heads and their classifiers are discarded at inference; only the main dual-path pool and its classifier are retained.
    
    \begin{table*}[width=\textwidth,cols=6,pos=!t]
    \caption{Comparison with state-of-the-art methods on Diving48. NA denotes not reported.}\label{tab:sota}
    \begin{tabular*}{\tblwidth}{@{} LCCCCC @{} }
    \toprule
    Model / Architecture & Backbone & Frames & Computational Cost (GFLOPs) & Top-1 Accuracy (\%) & Top-5 Accuracy (\%) \\
    \midrule
    TSN (baseline) \citep{tsn} & ResNet-50 & 16 & 33.0 & 79.0 & NA \\
    GST \citep{gst} & ResNet-50 & 16 & 58.4 & 78.9 & NA \\
    TSM \citep{tsm} & ResNet-50 & 16 & 65.0 & 83.2 & NA \\
    SlowFast (16$\times$8) \citep{slowfast} & ResNet-101 & 64+16 & 213.0 & 77.6 & NA \\
    TDN (Base) \citep{tdn} & ResNet-50 & 16 & 72.0 & 84.6 & NA \\
    TimeSformer-HR \citep{timesformer} & Transformer & 16 & 170.3 & 78.0 & NA \\
    TimeSformer-L \citep{timesformer} & Transformer & 96 & 2,380.0 & 81.0 & NA \\
    VideoSwin-B \citep{videoswin} & Transformer & 32 & 963.0 & 69.6 & 92.7 \\
    ViViT-L \citep{vivit} & Transformer & 32 & 7,248.0 & 80.6 & 92.7 \\
    SIFAR-B-12+ \citep{sifar} & Swin-B & 16 & 189.0 & 85.3 & 98.3 \\
    SIFAR-B-14+ \citep{sifar} & Swin-B & 16 & 263.0 & \textbf{87.3} & \textbf{98.8} \\
    \midrule
    \textbf{MamBOA (Ours, Image)} & VMamba-B \citep{vmamba} & 30 $\times$ 16 & 30 $\times$ 132.5\textsuperscript{a} & 85.02 & 98.53 \\
    \textbf{MamBOA (Ours, Video)} & MViT-V2-S \citep{mvitv2} & Full-video & 407.6 & 86.24 & 97.72 \\
    \bottomrule
    \end{tabular*}
    \begin{flushleft}
    \footnotesize
    \textsuperscript{a} MamBOA (Image-backbone) is evaluated using $10~\text{temporal views} \times 3~\text{spatial crops}$ multi-view inference. Several baseline models were not originally evaluated on Diving48 in their respective publications; their results on this benchmark are as reported in \citep{sifar} and \citep{dualpath}.
    \end{flushleft}
    \end{table*}
    
    \subsection{Comparison with State of the Art}
    Table \ref{tab:sota} compares MamBOA against published methods on Diving48. We group methods by pretraining strategy (image vs. video) and report the total inference FLOPs to enable a fair efficiency comparison. While the image-pretrained MamBOA variant features a high total GFLOPs count in Table \ref{tab:sota}, this overall footprint is a direct consequence of multiplying a single-view forward pass by the standard 30-view inference protocol ($10~\text{temporal} \times 3~\text{spatial}$) commonly used to evaluate image-pretrained models on video tasks. This multi-view inference bottleneck is what our clip-level video variant directly resolves by processing the full sequence in a single 407.6 GFLOP pass. To ensure a rigorous and equitable evaluation, our benchmark is strictly limited to foundational end-to-end architectures. Consequently, we exclude hybrid pipelines that rely on auxiliary frame-selection or heuristic filtering modules prior to the core network. Furthermore, all reported baselines exclusively utilize RGB frames as input, ensuring a controlled comparison devoid of optical flow or multi-modal computational overhead.

    Although the evaluation is conducted on a single benchmark, the generality of the proposed method is systematically validated along three independent architectural axes within that benchmark. First, the backbone-agnostic experiments (Table~\ref{tab:backbone_agnostic}) demonstrate consistent temporal modeling across three structurally distinct backbone families: CNN, Transformer, and Mamba. Second, the processing-block-agnostic experiments (Table~\ref{tab:processing_block_agnostic}) show that the BOA module retains its function regardless of the internal block family. Third, two independent pretraining regimes (image-pretrained frame-pair and video-pretrained clip-pair) confirm that the differential synthesis principle holds across temporal granularities. Taken together, these axes constitute generalization evidence of structural scope that is complementary to dataset diversity. This empirical breadth is further corroborated by mechanistic evidence: the temporal shuffling experiments (Section~\ref{sec:temporal_sensitivity}) confirm that the model's performance is causally grounded in temporal ordering rather than static appearance, the $\omega$ convergence analysis (Section~\ref{sec:temporal_sensitivity}) demonstrates that the differential extraction mechanism operates as theoretically intended without explicit inductive forcing, and the attention visualizations show that the dual-path pooling dynamically discovers discriminative moments rather than exploiting fixed positional biases.
    
    \subsection{Efficiency Analysis}
    The defining computational advantage of MamBOA lies in its architectural efficiency and strictly bounded operational overhead. Consequently, the primary computational footprint of the entire network is strictly dictated by the chosen base backbone, rather than the temporal modeling mechanism itself.

MamBOA is designed with computational efficiency in mind. The differential synthesis and dual-path pooling operate with a very low computational cost of merely $\sim$2.1 GFLOPs per feature pair (measured with ConvNeXt-V2 as the BOA processing block). The temporal head itself is computationally lightweight and contains only 34.63M parameters, which are loaded independently of the choice of spatial backbone.

Because this deep temporal alignment incurs only a minimal processing penalty, the overall efficiency of MamBOA dynamically scales with the chosen underlying network. This strict parameter isolation allows the architecture to process the entire video sequence in a single forward pass without the severe computational bottlenecks inherent to dense 3D networks, as detailed in Table \ref{tab:efficiency_flops}. To put this in concrete terms: VideoSwin-B and ViViT-L consume 963 and 7,248 GFLOPs respectively for a fixed 32-frame window, whereas MamBOA achieves full-video coverage at 407.6 GFLOPs total—of which only 12.6 GFLOPs ($2.1 \times 6$ pairs) are attributable to the temporal framework itself.

\begin{table}[pos=!htb]
\caption{Computational efficiency comparison. MamBOA isolates its temporal computational footprint from the spatial backbone, enabling full-sequence coverage at a fraction of the cost of dense 3D architectures.}\label{tab:efficiency_flops}
\begin{tabular*}{\tblwidth}{@{} LCC @{}}
\toprule
Method & Frames & Total FLOPs (G) \\
\midrule
VideoSwin-B \citep{videoswin} & 32 & 963 \\
ViViT-L \citep{vivit} & 32 & 7248 \\
\textbf{MamBOA (MViT-V2-S)} & \textbf{All Frames} & \textbf{407.6} (12.6\textsuperscript{a} + 395) \\
\bottomrule
\end{tabular*}
\begin{flushleft}
\footnotesize
\textsuperscript{a} The MamBOA architecture processes 6 temporal feature pairs in this configuration, requiring merely $\sim$2.1 GFLOPs per pair (measured with ConvNeXt-V2 as the BOA processing block; $2.1 \times 6 = 12.6$ GFLOPs total for the temporal framework). The remaining 395 GFLOPs are strictly native to the underlying MViTV2-S video-pretrained backbone.
\end{flushleft}
\end{table}

    \subsection{Backbone Agnostic Architecture}
    Our proposed MamBOA framework is inherently backbone-agnostic. Its temporal modeling mechanism can be seamlessly integrated with various spatial architectures. As demonstrated in Table \ref{tab:backbone_agnostic}, we evaluate the framework across different backbone families, including Convolutional Neural Networks (CNN), Transformers, and Mamba-based architectures, yielding competitive performance regardless of the underlying spatial backbone.

    \begin{table}[pos=!htb]
    \caption{Performance of MamBOA across different backbone families on Diving48.}\label{tab:backbone_agnostic}
    \begin{tabular*}{\tblwidth}{@{} LC @{}}
    \toprule
    Backbone & Top-1 Acc (\%) \\
    \midrule
    \textbf{CNN Family} & \\
    ConvNeXt-V2-B\textsuperscript{a} \cite{convnextv2} & 82.49 \\
    \midrule
    \textbf{Transformer Family} & \\
    Swin-B\textsuperscript{a} \cite{swin} & 78.33 \\
    \midrule
    \textbf{Mamba Family} & \\
    VMamba-B & 85.02 \\
    \bottomrule
    \end{tabular*}
    \begin{flushleft}
    \footnotesize
    \textsuperscript{a} Note that the objective of this analysis is to demonstrate structural compatibility (plug-and-play capability), not to establish state-of-the-art performance for every architecture. Swin-B and ConvNeXt-V2 were evaluated using hyperparameters and depths optimized for VMamba (specifically, depths $[2, 2, 6]$ for Swin-B and $[3, 3, 9]$ for ConvNeXt-V2), without any backbone-specific tuning.
    \end{flushleft}
    \end{table}
     
    \subsection{Processing Block Agnostic Design}
    To demonstrate that the processing block within the BOA module is independent of the backbone architecture, we evaluate the framework using the VMamba-B spatial backbone while substituting the internal processing block with modern blocks from three different architectural families. As shown in Table \ref{tab:processing_block_agnostic}, the framework maintains its robust temporal modeling capabilities regardless of the chosen processing block, confirming its modular design. Notably, optimal performance is achieved when pairing the SSM-based VMamba backbone with a CNN-based ConvNeXt block inside the BOA module, rather than the VSS block. This cross-family synergy empirically proves that the framework's performance stems from genuine architectural complementarity rather than an inherent bias toward homogeneous SSM structures.

    \begin{table}[pos=!htb]
    \caption{Performance of MamBOA on Diving48 with different processing blocks instantiated within the BOA module, using VMamba-B as the spatial backbone.}\label{tab:processing_block_agnostic}
    \begin{tabular*}{\tblwidth}{@{} LC @{}}
    \toprule
    Processing Block & Top-1 Acc (\%) \\
    \midrule
    \textbf{CNN Family} & \\
    ConvNeXt-V2 Block \cite{convnextv2} & 85.02 \\
    \midrule
    \textbf{Transformer Family} & \\
    Standard ViT (Vision Transformer) Block & 82.39 \\
    \midrule
    \textbf{Mamba Family} & \\
    VSS Block & 82.94 \\
    \bottomrule
    \end{tabular*}
    \end{table}

\begin{figure*}[!t]
\centering
\includegraphics[width=0.9\textwidth]{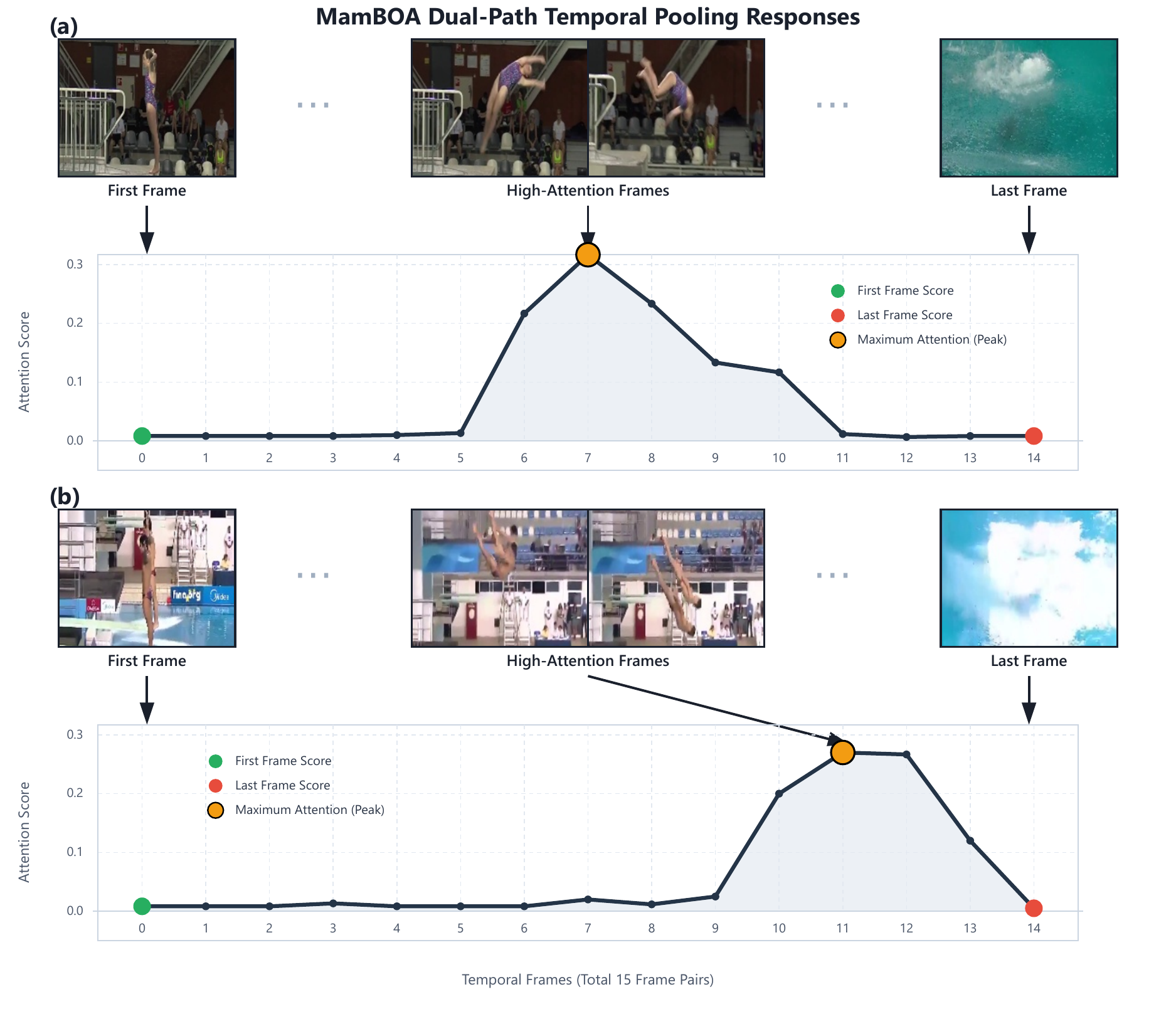}
\caption{Attention response curves ($\alpha_t$) from the dual-path pooling mechanism for two Diving48 instances. Peak attention dynamically aligns with the discriminative execution phases (e.g., pair 7 in (a) and pair 11 in (b)).}
\label{fig:temporal-pooling-attention}
\end{figure*} 

    \subsection{Ablation Studies on Network Dynamics}
    To investigate the contribution of each architectural component to the learned representations, we conduct a systematic network dynamics analysis, as summarized in Table~\ref{tab:ablation_components}. Each variant isolates a specific design decision by removing or replacing a single module while holding all other training conditions fixed. As shown in the table, the removal of the multi-scale scan or the learnable phase shift results in a performance drop of over 2.2\%, while omitting the dual-path pooling mechanism causes the most significant degradation (up to 3.7\%), demonstrating the critical role of each component in the framework's overall efficacy.

    \begin{table}[pos=!htb]
    \caption{Component ablation on Diving48 (image-pretrained). All variants share the same training configuration. $\Delta$ denotes the change relative to the full model.}\label{tab:ablation_components}
    \begin{tabular*}{\tblwidth}{@{} LCC @{}}
    \toprule
    Configuration & Top-1 (\%) & $\Delta$ \\
    \midrule
    Full model (MamBOA) & 85.02 & -- \\
    \midrule
    w/o Multi-Scale Scan (Only Single-Scale) & 82.79 & $-$2.23 \\
    w/o Learnable Phase Shift (no $\tau$) & 82.64 & $-$2.38 \\
    w/o FiLM \cite{film} (unidirectional) & 82.23 & $-$2.79 \\
    w/o dual-path pool (mean only) & 81.32 & $-$3.70 \\
    w/o dual-path pool (Attention only) & 81.68 & $-$3.34 \\
    w/o Attribute-specific heads & 83.20 & $-$1.82  \\
    w/o OHEM \cite{ohem} & 84.21 & $-$0.81 \\
    \bottomrule
    \end{tabular*}
    \end{table}
    
    \subsection{Temporal Sensitivity Analysis}\label{sec:temporal_sensitivity}

The foundational objective of the MamBOA framework is to explicitly capture and leverage temporal dynamics. In datasets like Diving48, where diverse action categories share nearly identical static visual contexts (e.g., diving boards and water splashes) and are distinguished solely by sequential motion variations (e.g., the exact number of aerial twists), structural temporal sensitivity is paramount. To rigorously quantify MamBOA's reliance on temporal ordering versus static appearance, we conduct targeted temporal shuffling experiments.

For the image-pretrained variant, randomly shuffling the sequence of input frames collapses the Top-1 accuracy to 48.48\% (a severe degradation from 85.02\%). This catastrophic drop confirms that the architecture does not overfit to spatial background cues. Instead, the Interleaved Temporal Phase-shift Synthesis (ITPS) pipeline inherently relies on strict chronological ordering to compute accurate differential macro-motion.

For the video-pretrained variant, the underlying MViT-V2-S backbone is already fundamentally motion-aware due to its Kinetics-400 pretraining, possessing a robust 32-frame local receptive field per clip. However, when we apply a clip-level shuffle—disrupting the global sequence order while fully preserving the local 32-frame motion dynamics within each clip—the accuracy still experiences a substantial drop to 72.30\% (down from 86.24\%). This significant degradation mathematically isolates and highlights MamBOA's critical architectural contribution: the framework successfully establishes and aggregates long-range, inter-clip temporal connections. It proves that the dual-path pooling mechanism actively models the holistic temporal logic of the entire video sequence, rather than merely relying on the backbone's localized motion fragments.

Furthermore, we observed that the learnable projection coefficients in the Motion Decoding Module ($\omega_1 Z_1 - \omega_0 Z_0$) consistently converge to approximately equal, positive values during training (e.g., $\omega_1 \approx 0.998, \omega_0 \approx 1.000$). This empirical finding is highly significant: it confirms that without any structural forcing or explicit inductive bias, the network autonomously discovers that a direct, equal-weighted subtraction is the optimal strategy to cancel static spatial interference and extract the differential motion signal.
    \subsection{Visualizing Temporal Pooling Dynamics}
Diving48 is a fine-grained benchmark in which all 48 action categories share highly similar visual contexts and are predominantly differentiated by the temporal composition of the aerial phase, including the sequence and number of somersaults and twists performed between takeoff and water entry. Although the pre-takeoff preparation and water-entry phases may contain auxiliary contextual information, they are comparatively less informative for distinguishing action categories. In contrast, the airborne execution window contains the richest category-specific motion patterns and therefore represents the primary source of discriminative temporal information.

We visualize the attention response curves ($\alpha_t$) of the dual-path pooling mechanism across 15 temporal frame pairs in Fig.~\ref{fig:temporal-pooling-attention}. The resulting curves exhibit a behaviorally interpretable pattern that aligns precisely with this data structure. During the pre-takeoff preparation and the post-entry phases — segments that are visually stereotypical and category-invariant — the attention weights remain consistently suppressed. In sharp contrast, the attention response rises sharply and peaks precisely during the aerial execution window, where the discriminative somersault and twist dynamics unfold. For instance, the attention peak accurately aligns with a mid-sequence aerial execution in Fig.~\ref{fig:temporal-pooling-attention}a (pair 7) and adapts to a late-sequence rotation in Fig.~\ref{fig:temporal-pooling-attention}b (pair 11). This structure did not arise from any explicit supervision over temporal localization — the model was trained only on category labels. The emergence of phase-selective attention therefore constitutes direct behavioral evidence that the dual-path pooling mechanism learns to suppress category-invariant context and selectively weight the temporally localized, category-discriminative motion events without being guided to do so.

    \subsection{Adaptive Stride (For Video Backbone)}
    We analyze the impact of the number of temporal feature pairs on computational complexity and recognition accuracy. While an increased number of feature pairs incurs a higher computational cost, it provides a crucial boost in performance. As shown in Table \ref{tab:adaptive_stride}, increasing the number of pairs from 1 to 6 yields consistently higher accuracy.

    Assuming an average video in Diving48 contains approximately 150--220 frames, a single temporal clip from the video backbone covers a $16 \times 2$ receptive field (32 frames). When 6 pairs (generated from 7 overlapping clips) are utilized, the model achieves near-complete coverage of the entire video, capturing the full action sequence. Conversely, configurations with 1 or 3 pairs only observe a partial fraction of the video, depriving the network of the holistic temporal context necessary for accurate recognition.
\begin{table}[pos=!htb]
    \centering
    \caption{Effect of the number of clips and feature pairs on Top-1 accuracy.}\label{tab:adaptive_stride}
    \begin{tabular}{ccc}
    \toprule
    Clips & Pair Features & Top-1 Acc (\%) \\
    \midrule
    2 & 1 & 45.28 \\
    4 & 3 & 76.90 \\
    7 & 6 & 86.24 \\
    \bottomrule
    \end{tabular}
    \end{table}

    \section{Discussion and Limitations}

\paragraph{What the evidence establishes.}
Taken together, the experimental results support a specific causal chain. Temporal shuffling eliminates chronology while preserving all spatial content and collapses accuracy in both variants, ruling out appearance-based shortcuts. The convergence of the MDM coefficients ($\omega_1 \approx \omega_0 \approx 1$) confirms that the decoding stage exploits precisely the cancellation structure that motivated its design. The component ablations show that no single module carries the architecture alone, and the attention visualizations confirm that temporal aggregation responds to content rather than position — the behavior the dual-path formulation was constructed to produce.

\paragraph{Cross-family complementarity.}
An initially counter-intuitive finding deserves comment: the best
processing block inside the BOA module is not the architecturally
homogeneous VSS block but the CNN-based ConvNeXt-V2 block
(Table~\ref{tab:processing_block_agnostic}). We interpret this as a
division of inductive labor. The selective scan stages of MamBOA already
supply global, content-dependent sequential mixing; what the post-scan
refinement stage benefits from most is a complementary inductive prior
rather than a second global mixing operator of the same family. Notably,
the distinguishing structural component of ConvNeXt-V2 relative to the
VSS block is not depthwise spatial convolution — which both architectures
share — but the Global Response Normalization (GRN) layer, which
performs cross-channel feature competition by normalizing each channel's
activation against the global $\ell_2$ norm of all channels. We
hypothesize that this cross-channel competitive suppression is
particularly beneficial in the post-scan context: it can selectively
amplify the most salient differential motion channels while suppressing
redundant or noisy ones, providing a form of channel-wise sparsification
that global sequential mixing alone cannot achieve. A similar asymmetry appears in the
backbone-agnostic results (Table~\ref{tab:backbone_agnostic}): the
framework remains functional across CNN, Transformer, and Mamba
backbones, but the accuracy spread between them is non-trivial. We
emphasize that these transfer experiments reuse the hyperparameters and
truncated depths optimized for VMamba without any backbone-specific
tuning; the results should therefore be read as evidence of structural
compatibility, not as the performance ceiling of each backbone family.
Characterizing which properties of a spatial representation make it most
amenable to differential synthesis — feature smoothness, effective
receptive field, or pretraining objective — is an open question raised,
but not answered, by these experiments.

\paragraph{Limitations.}
Several limitations bound the scope of the present study, and we state
them explicitly. First, all quantitative conclusions are drawn from a
single benchmark. As argued in
the evaluation protocol, Diving48 was selected because its shared
visual environment structurally eliminates spatial shortcuts, maximizing
the causal attribution of performance to temporal modeling; this
controlled depth, however, is purchased at the cost of breadth. The
architectural-axis experiments (backbone families, processing blocks, and
pretraining regimes) provide generalization evidence that is orthogonal
to dataset diversity, but they do not substitute for it: whether the
demonstrated mechanisms transfer to domains with different motion
statistics — object-centric interactions, egocentric viewpoints, or
multi-actor scenes — remains an empirical question that we leave to
future work. Second, the differential formulation is strictly pairwise.
Motion is synthesized from two temporal positions at a time, and
higher-order temporal patterns spanning multiple transitions are
delegated to the downstream pooling stage rather than captured within the
recurrence itself; phenomena defined by acceleration or rhythm, rather
than displacement, are therefore modeled only indirectly. Third, the
multi-scale group size $g=3$ is justified by the structural boundary
analysis of Eqs.~\ref{eq:pre_transition}--\ref{eq:multi_scan} rather than
by an exhaustive empirical sweep; while the analysis constrains the
plausible range, intermediate values were not exhaustively enumerated due
to computational budget. Fourth, the LPSA stage corrects the dominant
component of the receptive-field misalignment through a 1D phase shift;
a full 2D alignment operator could in principle capture residual
vertical offsets, at the cost of additional parameters and a more complex
frequency-domain formulation. Fifth, the image-pretrained variant
inherits the 30-view inference protocol standard for image backbones on
video tasks, and its total inference cost is therefore dominated by view
replication; although the clip-level variant resolves this bottleneck,
single-pass inference for image-pretrained backbones remains an open
efficiency problem. None of these limitations undermines the central
claim — that a selective recurrence over an interleaved sequence can
serve as a differential motion synthesizer — but they delineate the
conditions under which that claim has, and has not yet, been tested.

\section{Conclusion}
 
In this work, we presented MamBOA, a backbone-agnostic temporal framework
that rethinks motion extraction for fine-grained video understanding.
Rather than relying on computationally heavy 3D dense operators or rigid,
hand-crafted geometric subtraction, MamBOA demonstrates that differential
motion can be synthesized directly from the structural arrangement of the
input: by interleaving consecutive spatial feature maps into a single
alternating sequence, the selective state-space recurrence ($S6$) is
structurally driven to build a joint encoding of both temporal positions,
from which a learnable asymmetric projection decodes the motion signal.
On Diving48, this principle yields 85.02\% Top-1 accuracy with an
image-pretrained backbone and 86.24\% with a video-pretrained backbone
operating over the full video sequence, while the temporal head itself
adds only $\sim$2.1 GFLOPs per feature pair. Beyond raw accuracy, the
temporal shuffling experiments, the convergence of the decoding
coefficients to the theoretically predicted equal-weighted subtraction,
and the content-driven behavior of the dual-path attention jointly
confirm that the mechanism operates as designed rather than exploiting
appearance shortcuts. Furthermore, the framework establishes that
state-space differential synthesis can interface with disparate
architectural families — including Vision Transformers and standard
CNNs — acting as a modular temporal head whose cost is decoupled from
the backbone footprint.
 
Future work will extend MamBOA to long-form, untrimmed video — most
concretely, through temporal action segmentation. The clip-level pair
features developed here are naturally suited to this task: within a
single action, consecutive clip representations are similar and the
differential encoding remains small, whereas at an action boundary the
two clips encode different activities and the differential signal peaks.
The magnitude of the decoded motion feature thus behaves as an intrinsic
boundary indicator, providing exactly the transition-sensitive signal
that segmentation requires. Under this formulation, a sliding window of
consecutive pair features would yield a structured sequence of
short-range action transitions, from which a lightweight sequence model
could infer segment boundaries and class assignments at clip-stride
temporal granularity. Realizing this direction will require replacing the
global dual-path pooling with a localized, windowed variant that
preserves per-segment outputs rather than collapsing the sequence into a
single summary, and adapting the hierarchical loss formulation to
segment-level supervision. In parallel, we aim to broaden the empirical
scope of the framework along two axes: validation on additional
fine-grained benchmarks with different motion statistics, and a
systematic characterization of which spatial-prior properties make a
backbone most amenable to state-space differential synthesis.

\section*{Reproducibility and Code Availability}
To support the reproducibility of this work, the source code and configuration files required to implement MamBOA are publicly available at: \texttt{https://github.com/BOA-clk/MamBOA}
\section*{Acknowledgements}

The author has no acknowledgments to declare.

\printcredits

    \end{document}